\pdfoutput=1

\documentclass[11pt]{article}

\usepackage[final]{ACL2023}

\usepackage{times}
\usepackage{latexsym}

\usepackage{times}
\usepackage{latexsym}
\usepackage[algoruled,ruled,vlined,noend]{algorithm2e}
\usepackage{amsmath}
\usepackage{times}
\usepackage{latexsym}
\usepackage{pgfplots}
\usepackage{graphicx}
\usepackage{enumitem}
\usepackage{tikz}
\usepackage{diagbox}
\usepackage{tkz-tab}
\usepackage{caption}
\usepackage{booktabs}
\usepackage{latexsym}
\usepackage{amssymb}
\usepackage{amsmath}
\usepackage{subcaption}
\usepackage{natbib}
\usepackage{blindtext}
\usepackage{url}
\usepackage{color}

\usepackage{url}
\usepackage{hyperref}    
\usepackage{cleveref}
\SetAlFnt{\small}
\SetAlCapFnt{\small}
\SetAlCapNameFnt{\small}
\usepackage{ntheorem}

\usepackage{booktabs, tabularx}
\usepackage{tfrupee}  
\usepackage{caption}

\usepackage{epigraph}

\usepackage[T1]{fontenc}

\usepackage[utf8]{inputenc}

\usepackage{microtype}

\usepackage{inconsolata}
\title{Dissecting In-Context Learning of Translations in GPTs}

\author{Vikas Raunak \qquad Hany Hassan Awadalla \qquad Arul Menezes\\\\
Microsoft Azure AI \\
Redmond, Washington \\
\texttt{\{viraunak,hanyh,arulm\}@microsoft.com}}

\begin{document}
\maketitle
\begin{abstract}

Most of the recent work in leveraging Large Language Models (LLMs) such as GPT-3 for Machine Translation (MT) has focused on selecting the few-shot samples for prompting. In this work, we try to better understand the role of demonstration attributes for the in-context learning of translations through perturbations of high-quality, in-domain demonstrations. We find that asymmetric perturbation of the source-target mappings yield vastly different results. We show that the perturbation of the source side has surprisingly little impact, while target perturbation can drastically reduce translation quality, suggesting that it is the output text distribution that provides the most important learning signal during in-context learning of translations. We propose a method named Zero-Shot-Context to add this signal automatically in Zero-Shot prompting. We demonstrate that it improves upon the zero-shot translation performance of GPT-3, even making it competitive with few-shot prompted translations.

\end{abstract}

\section{Introduction}

Recent work has put into question the importance of the correctness of demonstrations for prompting in Large Language Models (LLMs) \cite{min2022rethinking}. One key conjecture is that the latent zero-shot capabilities of LLMs might be considerably higher than their observed zero-shot capabilities for a range of tasks \cite{min2022rethinking, kojima-zero-shot}. One way to elicit higher zero-shot performance is to qualify the role of demonstration attributes towards task performance and then simulate such in-context learning signals in a zero-shot manner. However, realizing this goal hinges on explicitly dissecting the role of various demonstration attributes (format, inputs, outputs, input-output mapping) towards task performance within few-shot in-context learning. In this work, we explore these questions for the task of Machine Translation (MT). Our line of inquiry is orthogonal to finding the most useful samples for few shot learning, a topic that has received considerable attention for eliciting better translations from LLMs \cite{mt_incontext_1, mt_incontext_2}. Our contributions are:
\begin{enumerate}
    \item We explore the role of demonstration attributes within in-context learning of translations in the GPT family of LLMs, through perturbations of the input-output (source-target) mappings. We show that the target text distribution is the most important factor in demonstrations, while the source text distribution provides an inconsequential learning signal. 
    \item Based on our findings, we propose Zero-Shot-Context prompting, which tries to automatically provide the learning signal corresponding to the target text distribution without any source-target examples. This greatly improves GPT-3's zero-shot performance, even making it competitive with few-shot prompting.
\end{enumerate}

\begin{table*}[ht]
    \begin{tabularx}{\linewidth}{ X X X X X }
        \toprule
    \textbf{Ground Truth} & \textbf{Shuffled Targets} & \textbf{Jumbled Source} & \textbf{Jumbled Target} & \textbf{Reversed Target}  \\
        \midrule
\colorbox{lime}{English: A B C} \newline \colorbox{lime}{German: D E F} \newline \colorbox{lime}{English: U V W} \newline \colorbox{lime}{German: X Y Z} & \colorbox{lime}{English: A B C} \newline \colorbox{pink}{German: X Y Z} \newline \colorbox{lime}{English: U V W} \newline \colorbox{pink}{German: D E F} & \colorbox{pink}{English: B A C} \newline \colorbox{lime}{German: D E F} \newline \colorbox{pink}{English: U W V} \newline \colorbox{lime}{German: X Y Z} & \colorbox{lime}{English: A B C} \newline \colorbox{pink}{German: E D F} \newline \colorbox{lime}{English: U V W} \newline \colorbox{pink}{German: Y Z X} & \colorbox{lime}{English: A B C} \newline \colorbox{pink}{German: F E D} \newline \colorbox{lime}{English: U V W} \newline \colorbox{pink}{German: Z Y X} \\
        \bottomrule
    \end{tabularx}
    \caption{Perturbations Applied: The four types of perturbations (shown here as applied on abstract source-target example sequences) manipulate the demonstration attributes differently. For example, while Jumbled Source and Jumbled Target both corrupt the source-target mapping, they modify different learning signals in in-context learning.}
    \label{tab:table1_perturbations}
\end{table*}

\section{Related Work}
Our work is related to two key themes, namely prompting LLMs for translation and analysis of in-context learning in LLMs. In this section, we situate our work within these two themes.

\paragraph{LLM Prompting for MT:} LLMs have achieved close to the state-of-the-art translation performance under few-shot prompting \cite{hendy2023good, cite_1}. Most of the work for prompting in MT has focused on selecting the training or development instances to be used as examples during prompting. \citet{mt_incontext_1} experiment on PaLM \cite{chowdhery2022palm} and find that quality of examples is the most important factor in few-shot prompting performance. \citet{mt_incontext_2} experiment with XGLM \cite{xglm} and report that translation quality and the domain of the examples are consequential. Our work builds on these with a different aim, in that we do not explore selecting the examples, rather apply perturbations on high-quality, in-domain examples to better qualify the role of certain demonstration attributes for in-context learning of translations.

\paragraph{Analyzing In-Context Learning:} Theoretical and empirical investigation of in-context learning is an ongoing research endeavor \cite{icl_bayesian_inference, icl_gradient_descent, icl_akyurek2022learning, icl_meta_gradient}. \citet{min2022rethinking} demonstrate that label correctness in demonstrations is of limited importance for open-set classification tasks, while \citet{kimetal} show that negated labels do matter. Our experiments differ from these works both on the choice of the task (translation, which has an exponential output space) as well as on the types of perturbations applied to the demonstrations.

\section{The Role of Demonstration Attributes}
\label{sec:section3}
To produce outputs for a specific task, LLMs are typically prompted with demonstrations (input-output examples pertaining to the specific task) appended with the test input. Similar to \citet{min2022rethinking}, we posit that there exist four aspects of demonstrations of the translation task that provide a learning signal: the input-output mapping, the input text distribution, the output text distribution and the format. In this section, we conduct an empirical investigation on how LLMs such as GPT-3 leverage the demonstrations provided to them for the task of translation by perturbing the input-output (source-target) mappings provided during prompting. Through these experiments, we hope to compare the importance of three key demonstration attributes -- the input text distribution, the output text distribution and their mapping for translation.


\paragraph{Models:} In this section, we mainly report results for text-davinci-002 \footnote{LLMs: \url{https://beta.openai.com/docs/models/}}, one of the most capable LLM models publically accessible \cite{helm}. We also investigate the veracity of our observations with text-davinci-001 and text-curie-001, two prior LLM versions in the GPT family as well as the more recent text-davinci-003.

\paragraph{Datasets:} We experiment with the WMT'21 News Translation task datasets \cite{wmt-2021-machine}, for the following four language pairs: English-German (En-De), German-English (De-En), English-Russian (En-Ru) and Russian-English (Ru-En). On each of these datasets text-davinci-002 achieves highly competitive performance with the WMT-21 winning NMT model \cite{facebook_wmt21}, with eight demonstrations ($k=8$ in \textit{k}-shot prompting). We list the full test set performance with text-davinci-002 and text-davinci-003 for $k=8$ in Table \ref{tab:table4}, while the perturbation experiments are reported on 100 random samples from the test sets in each case.

\begin{table}[!htbp]
\centering
\scalebox{0.8}{
\begin{tabular}{l|c|c|c|c}
\hline
Method & \textbf{En-De} & \textbf{De-En} & \textbf{Ru-En}& \textbf{En-Ru} \\ \hline
Facebook-WMT-21  & 39.36 & 39.88 & 35.25 & 46.41 \\ 
davinci-002 (k=8)  & 39.57 & 40.28 & 35.67 & 39.06 \\ 
davinci-003 (k=8)  & 40.31 & 41.31 & 36.03 & 41.82 \\ \hline
\end{tabular}}
\caption{COMET-QE Scores on WMT-21 Test Sets: Both the translations from the WMT-21 winning system \cite{facebook_wmt21} as well as the GPT translations were obtained through greedy decoding.}
\label{tab:table4}
\vspace{-0.4em}
\end{table}

\paragraph{Prompt Details:} \citet{mt_incontext_1} report than the choice of the format is inconsequential for few-shot prompting on the translation task. As such, we use the standard prompt used for MT in prior works, namely [Source]: ABC (\textbackslash n) [Target]: DEF, where Source (e.g., English) and Target (e.g., German) represent the language names. Further, we use high-quality, in-domain sentence pairs sampled from the development set for few-shot prompting.

\paragraph{Evaluation:} To minimize reference-bias in evaluation, which has been shown to be detrimental in estimating the LLM output quality in related sequence transduction tasks \cite{goyal2022news, garcia, raunak-etal-2023-gpts}, we make use of a state-of-the-art Quality Estimation \cite{qe_fomicheva} metric named COMET-QE \cite{comet} for quality evaluation. Further, one caveat of using the reference-free metric is that it allocates high scores to a translation if it is in the same language as the source sentence, i.e. it doesn't penalize copy errors in translation. To mitigate this evaluation shortcoming, we use a language-id classifier \cite{fastext-langid} and set the translation to empty if the translation is produced in the same language as the source.

\paragraph{Experiment 1:} We apply four perturbations to the demonstrations used for prompting. Table \ref{tab:table1_perturbations} enumerates the four perturbations with abstract source-target sequences: Shuffled Targets (ST) randomizes the mappings between the source and targets in the prompt, Jumbled Source (JS) randomizes the position of the words in the source sentences, Jumbled Ref (JT) randomizes the positions of the words in the target sentences and Reversed Ref (RT) reverses the order of the words in the target sentence. Among these perturbations, ST impacts both the input and output spaces symmetrically, while the other perturbations (JS, JT and RT) perturb only one of the input/output spaces.  

\begin{figure}[ht]
\centering
\begin{subfigure}[b]{0.48\textwidth}
\centering
\includegraphics[width=\textwidth]{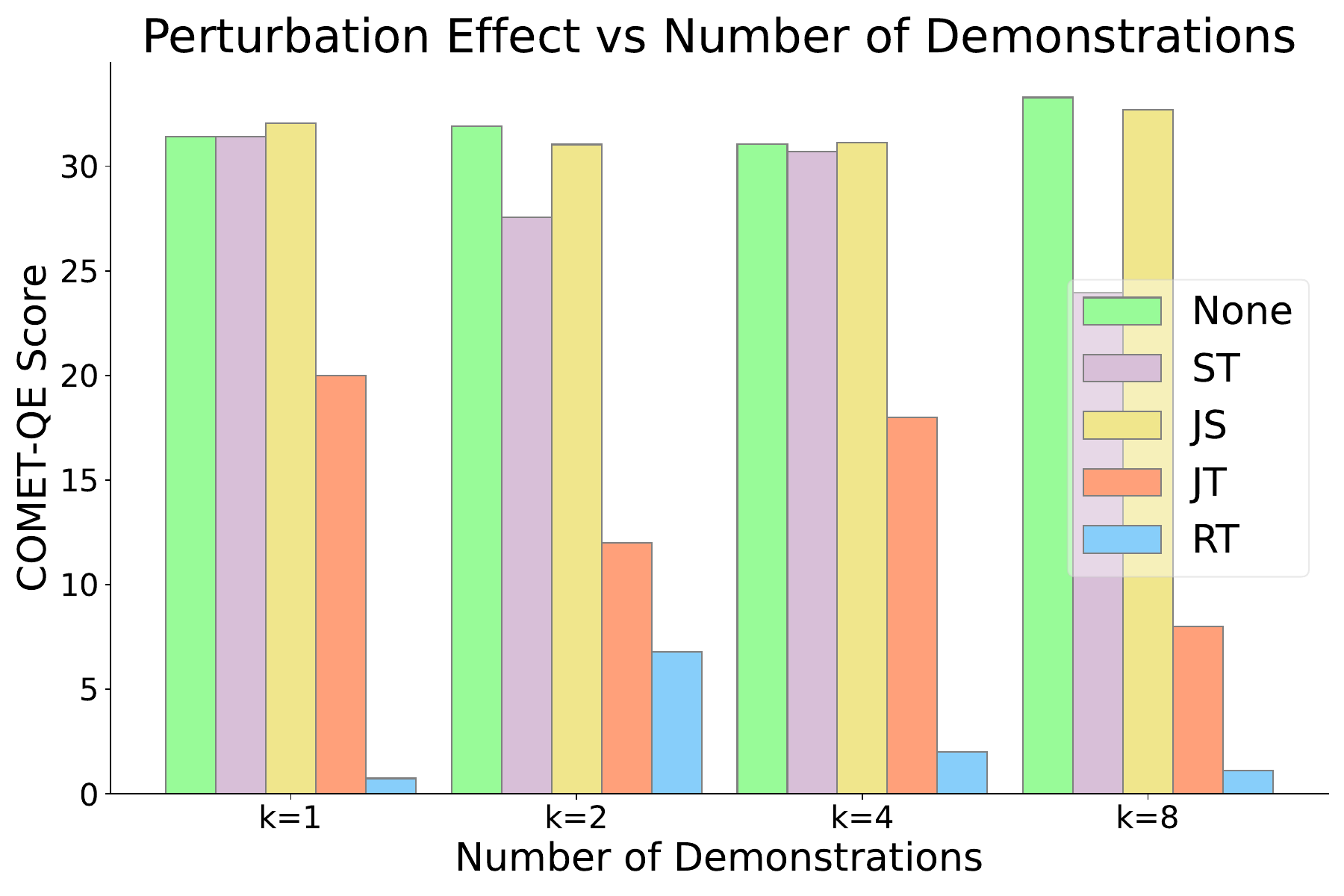} 
\end{subfigure}
\caption{Perturbing the demonstrations for WMT-21 English-German test set. Source and Target perturbations have asymmetric effects despite the input-output mapping getting severely damaged in both cases.}
\label{fig:perturbation_ende}
\end{figure}

\paragraph{Results:} The results of applying these perturbations on En-De are presented in Figure \ref{fig:perturbation_ende}, across different number of demonstrations ($k=1,2,4,8$). The results show that while ST and JT both significantly disrupt the source-target mappings in the demonstrations, they have greatly different impact. Translation quality declines by a large value for JT, an effect that becomes larger with increasing $k$, e.g., for JT perturbation at $k=8$, the translation quality is considerably worse. On the other hand, JS produces very little to no effect on the quality of translations. Further, owing to the nature of the perturbation ST becomes more disruptive at higher values of $k$, while yielding no impact for $k=1$.

 \paragraph{Experiment 2:} We repeat the same experiment as above (Experiment 1) with four different language pairs from WMT-21 and text-davinci-002.

\begin{figure}[ht]
\centering
\begin{subfigure}[b]{0.48\textwidth}
\centering
\includegraphics[width=\textwidth]{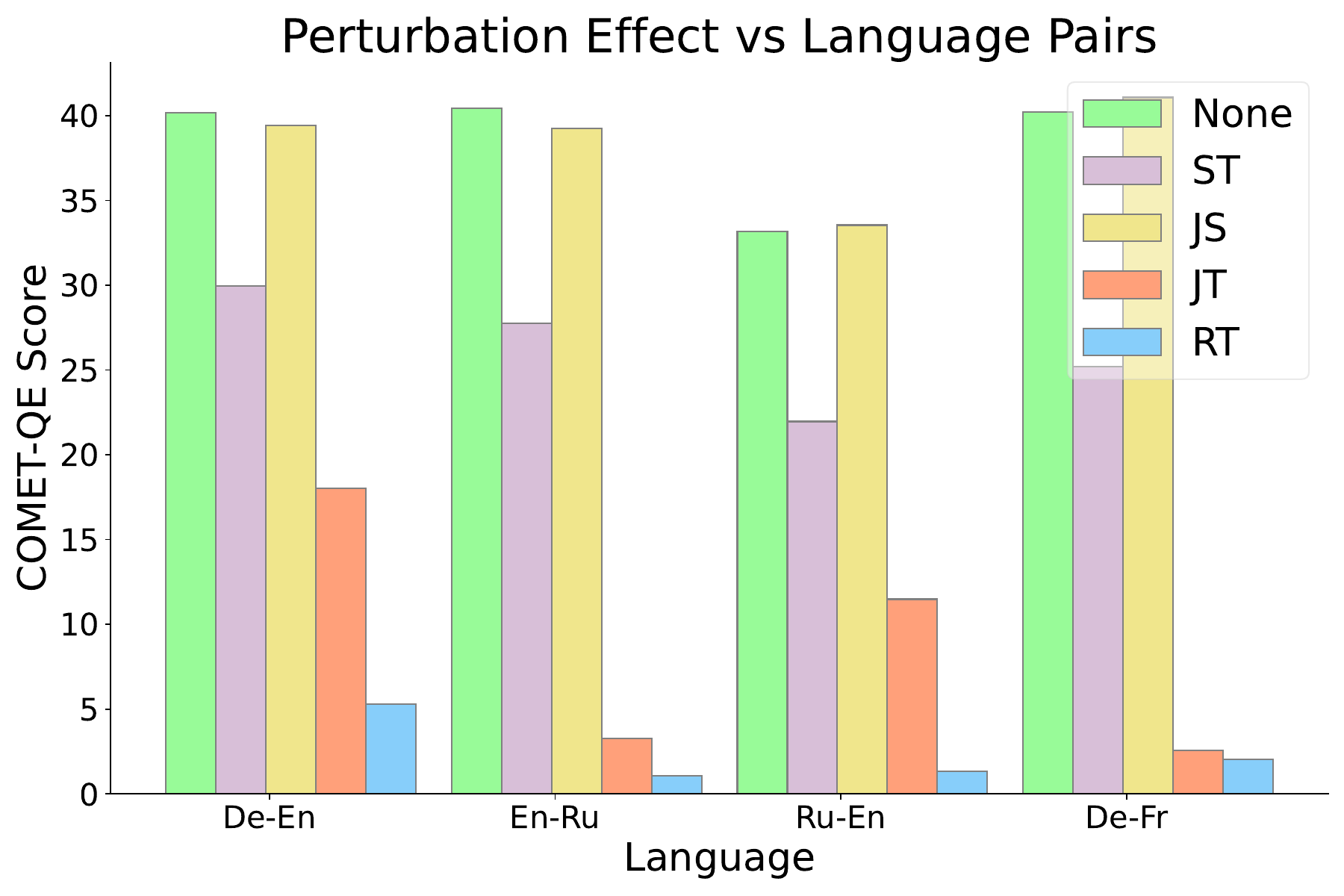} 
\end{subfigure}
\caption{Perturbation effects across different WMT'21 language pairs for text-davinci-002, under few-shot prompting with \textit{k}=8. The asymmetric effect of source and target perturbation holds true throughout the pairs.}
\label{fig:perturbation_datasets}
\end{figure}

\paragraph{Results:} The results are reported in Figure \ref{fig:perturbation_datasets}. We find that the trends are similar to the first experiment (Figure \ref{fig:perturbation_ende}). Across the language pairs, JS and JT have asymmetric impact on translation quality, showing that in each case the critical learning signal arrives from the target text distribution, while the source text distribution is an inconsequential factor with respect to the output translation quality.

\begin{figure}[ht]
\centering
\begin{subfigure}[b]{0.48\textwidth}
\centering
\includegraphics[width=\textwidth]{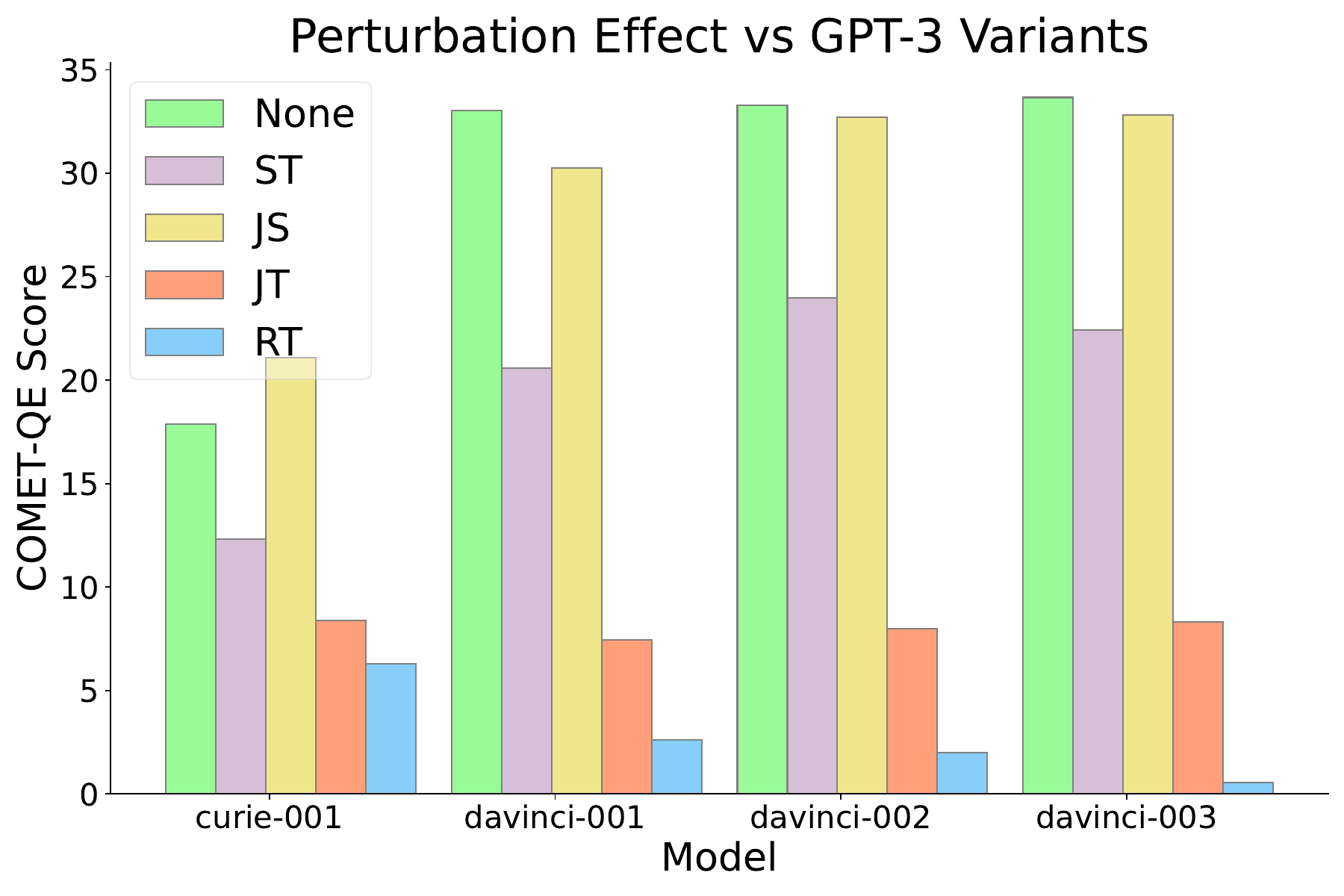} 
\end{subfigure}
\caption{Perturbation effect across GPT-3 model variants for the WMT-21 English-German test set. The asymmetric effect of source and target perturbation holds across different models, suggesting that this is a stable trait of the in-context learning mechanism.}
\label{fig:perturbation_models}
\end{figure}

\paragraph{Experiment 3:} We repeat Experiment 2, by keeping the language pair fixed to En-De and varying the LLMs. We report results in Figure \ref{fig:perturbation_models} for three other models from the GPT family, namely text-curie-001, text-davinci-002 and text-davinci-003.

\paragraph{Results:} We find that across different models, JS and JT have asymmetric impact on the translation quality, consistent with the prior two experiments.

\paragraph{Analysis:} Compared to \citet{min2022rethinking}, wherein the randomization of the input-output mappings in the demonstrations leads to \textit{better} performance than no demonstrations (zero-shot prompting) for open-set classification tasks, our results are quite different. We find that \textit{depending on the type of perturbation}, in-context translation learning results can be vastly different \textit{even when all the perturbations break the correct input-output mapping}. For some perturbations (e.g., JT and RT) the translation quality is much worse than zero-shot. To reconcile these results, we hypothesize that the difference arises from the increased complexity of the auto-regressive search in the case of translation, i.e., a clear specification of the output space in the demonstrations becomes much more critical to constrain the search space. 

Further, the ST results in Figures \ref{fig:perturbation_datasets} \& \ref{fig:perturbation_models} show that source-target mapping is also a critical demonstration attribute, a fact consistent with prior results emphasizing the importance of example quality \cite{mt_incontext_1, mt_incontext_2}. However, we show that it is not the primary learning signal in in-context learning of translations and even therein the source word order matters for little, suggesting that only an approximation of the input text distribution is sufficient for effective in-context learning.

\paragraph{Generality of Our Findings:} We also conduct experiments on gpt-3.5-turbo-instruct and gpt-3.5-turbo-instruct-0914, two of the more recent LLMs in the GPT family. With gpt-3.5-turbo-instruct on En-De, no perturbation (\textit{None} in the plots) obtains a COMET-QE score of 34.21, the JS perturbation a score of 35.20 and the JT perturbation obtains a score of 25.45. Similarly, with gpt-3.5-turbo-instruct-0914 on En-De, no perturbation obtains a COMET-QE score of 33.64, the JS perturbation a score of 34.35 and the JT perturbation obtains a score of 24.42. This observed behavior is agnostic to the choice of the MT quality metric as well: with COMET-KIWI (the state-of-the-art QE metric in the WMT-22 Quality Estimation Shared Task \cite{rei-etal-2022-cometkiwi}), no perturbation (None in the plots) with gpt-3.5-turbo-instruct obtains a score of 83.75, the JS perturbation a score of 83.94 and the JT perturbation obtains a score of 73.26. Similarly, with COMET-KIWI gpt-3.5-turbo-instruct-0914 obtains a score of 83.94, the JS perturbation a score of 83.85 and the JT perturbation obtains a score of 72.72. These results point to the robustness of our findings. 

\paragraph{Implications:} Our findings suggest that the data representing the output space might be the most important attribute in demonstrations for in-context learning of translations. Besides suggesting an in-built robustness towards perturbations on the source side, this result points to interesting exploratory directions for data selection for prompting, e.g., that target-original data might be more useful as demonstration examples than source-original. We leave such questions to future work.

\section{Zero-Shot-Context for Translation}
Previously, we demonstrated that the most important demonstration attribute for in-context learning of translations is the output text distribution. In this section, we present a method of providing this learning signal in a zero-shot manner. Our experiment here represents an inverse of experiments in section \ref{sec:section3}, i.e., here we \textit{add a useful learning signal to zero-shot prompting}, rather removing learning signals from few-shot prompting to gauge their importance. We present a method named `Zero-Shot-Context' and show that it greatly improves upon zero-shot performance for GPT-3, eliciting performance competitive even with few-shot prompting. Note that this method is one \textit{example} of adding a particular signal in zero-shot prompting and that there could be multiple ways to add such a signal to bolster zero-shot performance including direct instruction finetuning on the translation task. However, we leave a thorough analysis of improving zero-shot translation performance by adding relevant signals from demonstrations to future work and focus only exploring on our key hypothesis.


\begin{figure}[ht]
\centering
\begin{subfigure}[b]{0.49\textwidth}
\centering
\includegraphics[width=\textwidth]{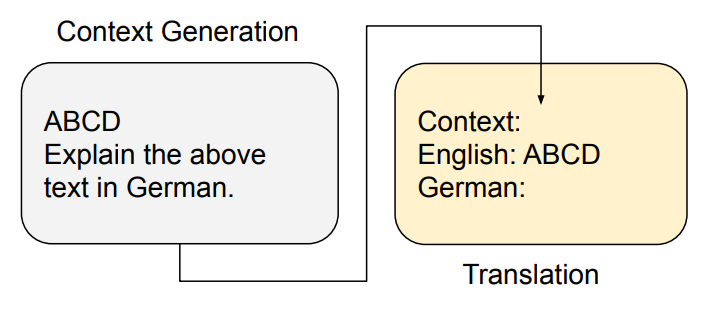} 
\end{subfigure}
\caption{Schematic for Zero-Shot-Context: The Context Generation step provides an automatic learning signal to the LLM about the output text distribution, simulating the most important demonstration attribute.}
\label{fig:schematic_zscot}
\end{figure}

\paragraph{Proposed Method:} We propose a new zero-shot prompting method named Zero-Shot-Context (Figure \ref{fig:schematic_zscot}), which auto-generates the output space specification learning signal from the LLM itself (the \textit{Context}) and uses it to condition the translation. 

\begin{table}[!htbp]
\centering
\scalebox{0.8}{
\begin{tabular}{l|c|c|c|c}
\hline
Method & \textbf{CQE$\uparrow$} & \textbf{BLEU$\uparrow$} & \textbf{ChrF$\uparrow$}& \textbf{TER$\downarrow$} \\ \hline
\textit{Zero-Shot}   & \textit{32.29} & 22.6 & 54.3  & 71.4 \\ 
\textit{Zero-Shot-Context}  & \textit{37.65} & 23.1 & 55.4 & 68.5 \\ 
Few Shot (k=1)  & 39.92 & 22.4 & 54.1 & 71.8 \\ 
Few Shot (k=2)  & 39.04 & 24.7 & 56.6 & 64.8 \\ 
Few Shot (k=4)  & 40.36 & 24.0 & 55.7 & 65.4 \\ \hline
\end{tabular}}
\caption{Zero-Shot-Context vs Baselines on WMT-21 En-De: Zero-Shot-Context greatly improves upon Zero-Shot Translations, gaining +5 QE points in quality.}
\label{tab:table2}
\end{table}

\begin{table}[!htbp]
\centering
\scalebox{0.8}{
\begin{tabular}{l|c|c|c|c}
\hline
Method & \textbf{CQE$\uparrow$} & \textbf{BLEU$\uparrow$} & \textbf{ChrF$\uparrow$}& \textbf{TER$\downarrow$} \\ \hline
\textit{Zero-Shot}   &  \textit{35.39}  & 19.8 & 49.4 & 74.3 \\ 
\textit{Zero-Shot-Context}   & \textit{40.67} & 18.8 & 48.7 & 75.6 \\
Few Shot (k=1)  & 37.92  & 20.5 & 50.1 & 72.3 \\ 
Few Shot (k=2)  & 39.35  & 19.3 & 50.0 & 72.7 \\ 
Few Shot (k=4)  & 39.25  & 20.2 & 50.1 & 72.3 \\ \hline
\end{tabular}}
\caption{Zero-Shot-Context vs Baselines on WMT-21 En-Ru: Zero-Shot-Context greatly improves upon Zero-Shot and is even competitive with few-shot translations.}
\label{tab:table3}
\end{table}

\paragraph{Experiment and Results:} In Table \ref{tab:table2} we compare Zero-Shot-Context with Zero-Shot prompting, as well as few-shot prompting (for $k$=1, 2, 4) with high-quality, in-domain examples sampled from the development set, on En-De WMT-21 test set with text-davinci-002. The results show that Zero-Shot-Context greatly improves upon Zero-Shot translation quality as measured by COMET-QE (CQE). Note that the gains are not visible in reference-based evaluation with BLEU and ChrF and limitations of these metrics have been pointed out in the literature \cite{no_bleu}. Table \ref{tab:table3} presents a comparison on the WMT-21 En-Ru test set.

\paragraph{Ablation on Zero-Shot Context:} We consider the following experiment: we pick a random target-side sentence from the development set and replace the Context-Generation step's output with the random target-side sentence. Ideally, an in-domain, high-quality target-side sentence should also be able to provide a learning signal regarding the output text distribution. We find that this is indeed the case, and simply replacing the context generation step with the random target-side sentence also improves upon zero-shot performance, achieving 36.10 COMET-QE score for WMT-21 En-De test set and 37.86 COMET-QE score for WMT-21 En-Ru. However, these scores are lower than Zero-Shot-Context, suggesting that the contextual nature of Zero-Shot-Context is also important.

\paragraph{Further Analysis:} Our findings indicate that the latent zero-shot GPT-3 performance for translations could indeed be higher than currently reported and that it is possible to leverage \textit{direct computation} to improve LLM translation performance instead of manually retrieving or selecting examples. In particular, we showed that a simple addition of a signal pertaining to the output space improved the zero-shot performance of text-davinci-002, a useful step towards better zero-shot utilization of LLMs for translation. As pointed out in \citet{bawden-yvon-2023-investigating}, generating zero-shot translations often suffers from outputs in the wrong language and we find that Zero-Shot-Context considerably alleviates this, leading to better performance. However, further rigorous analysis of this phenomenon across different LLMs is hindered by the fact that we do not have access to the training or the instruction finetuning dataset used for the underlying state-of-the-art LLMs.

\section{Summary and Conclusions}
We analyzed the relative importance of demonstration attributes as learning signals within few-shot in-context learning of translations in LLMs from the GPT family. We demonstrated that the critical learning signal arrives from the output text distribution, followed by the input-output mapping, while the input text distribution matters for little. We use this finding to propose Zero-Shot-Context, a method that tries to automatically generate the critical learning signal. Zero-Shot-Context greatly improves upon zero-shot translation quality in GPT-3, further validating our findings. We hope that our work could serve as a useful contribution towards better understanding of in-context learning of translations in LLMs.

\section{Limitations}
Our work experiments with high-quality, in-domain examples for few-shot prompting. It is conceivable that perturbations could have different impacts on examples with varying quality. Also, while our proposed zero-shot method does not consume any manual examples, it suffers from the limitation that it involves two passes over a LLM. While this is mitigated by the method presented as an ablation, we believe that simpler methods to add the relevant demonstration signal could be derived by pre-computing the singular target-side context once for the entire test set, a proposal we didn't investigate.

\bibliography{anthology,custom}
\bibliographystyle{acl_natbib}

\end{document}